\documentclass[11pt]{article}
\PassOptionsToPackage{breaklinks}{hyperref}

\usepackage[hyperref]{emnlp-ijcnlp-2019}
\usepackage{times}
\usepackage{latexsym}

\usepackage{graphicx}
\usepackage{multirow}
\usepackage{multicol}
\usepackage{amssymb}
\usepackage{wasysym}
\usepackage{arydshln}

\aclfinalcopy

\newcommand{\Tab}[1]{Table~\ref{tab:#1}}
\newcommand{\Sec}[1]{Section~\ref{sec:#1}}

\newcommand{\winM}{$^{\bullet}$}

\newcommand{\unidir}[2]{{#1$\rightarrow$#2}}
\newcommand{\bidir}[2]{{#1$\leftrightarrow$#2}}

\newcommand{\affiA}{{$^{\dagger}$}}
\newcommand{\affiB}{{$^{\ddagger}$}}

\newcommand{\RNum}[1]{\uppercase\expandafter{\romannumeral #1\relax}}

\title{Exploiting Out-of-Domain Parallel Data through Multilingual Transfer Learning for Low-Resource Neural Machine Translation\thanks{\, The contents in this manuscript are identical to those in our formal publication at the 17th Machine Translation Summit, whereas the style is slightly modified.}}

\author{Aizhan Imankulova{\affiA}\quad
Raj Dabre{\affiB}\quad Atsushi Fujita{\affiB}\quad Kenji Imamura{\affiB}\\
{\affiA}Tokyo Metropolitan University\\
6-6 Asahigaoka, Hino, Tokyo 191-0065, Japan\\
{\tt imankulova-aizhan@ed.tmu.ac.jp}\\
{\affiB}National Institute of Information and Communications Technology\\
3-5 Hikaridai, Seika-cho, Soraku-gun, Kyoto, 619-0289, Japan\\
\{\tt raj.dabre, atsushi.fujita, kenji.imamura\}@nict.go.jp}

\date{}

\begin{document}
\maketitle
\begin{abstract}
\normalsize
This paper proposes a novel multilingual multistage fine-tuning approach for low-resource neural machine translation (NMT), taking a challenging Japanese--Russian pair for benchmarking.
Although there are many solutions for low-resource scenarios, such as multilingual NMT and back-translation,
we have empirically confirmed their limited success when restricted to in-domain data. We therefore propose to exploit out-of-domain data through transfer learning, by using it to first train a multilingual NMT model followed by multistage fine-tuning on in-domain parallel and back-translated pseudo-parallel data. Our approach, which combines domain adaptation, multilingualism, and back-translation, helps improve the translation quality by more than 3.7 BLEU points, over a strong baseline, for this extremely low-resource scenario.
\end{abstract}

\section{Introduction}
\label{sec:intro}

Neural machine translation (NMT) 
\cite{DBLP:journals/corr/ChoMGBSB14,DBLP:journals/corr/SutskeverVL14,DBLP:journals/corr/BahdanauCB14:original} has enabled end-to-end training of a translation system without needing to deal with word alignments, translation rules, and complicated decoding algorithms, which are the characteristics of phrase-based statistical machine translation (PBSMT) \cite{koehn-EtAl:2007:PosterDemo}.
Although NMT can be significantly better than PBSMT in resource-rich scenarios, PBSMT performs better in low-resource scenarios \cite{koehn-knowles:2017:NMT}. 
Only by exploiting cross-lingual transfer learning techniques \cite{DBLP:journals/corr/FiratCB16:original,DBLP:conf/emnlp/ZophYMK16:original,kocmi-bojar-2018-trivial}, can the NMT performance approach PBSMT performance in low-resource scenarios.

However, such methods usually require an NMT model trained on a resource-rich language pair like \bidir{French}{English} (parent), which is to be fine-tuned for a low-resource language pair like \bidir{Uzbek}{English} (child). On the other hand, multilingual approaches \cite{TACL1081} propose to train a single model to translate multiple language pairs. However, these approaches are effective only when the parent target or source language is relatively resource-rich like English (En).
Furthermore, the parents and children models should be trained on similar domains; otherwise, one has to take into account an additional problem of domain adaptation \cite{chu:2017:ACL:original}.

In this paper, we work on a linguistically distant and thus challenging language pair \bidir{Japanese}{Russian} (\bidir{Ja}{Ru}) which has only 12k lines of news domain parallel corpus and hence is extremely resource-poor. Furthermore, the amount of indirect in-domain parallel corpora, i.e., \bidir{Ja}{En} and \bidir{Ru}{En}, are also small. As we demonstrate in \Sec{baseline}, this severely limits the performance of prominent low-resource techniques, such as multilingual modeling, back-translation, and pivot-based PBSMT. To remedy this, we propose a novel multistage fine-tuning method for NMT that combines multilingual modeling \cite{TACL1081} and domain adaptation \cite{chu:2017:ACL:original}.

We have addressed two important research questions (RQs) in the context of extremely low-resource machine translation (MT) and our explorations have derived rational contributions (CTs) as follows:

\begin{description}\itemsep=0mm
\item [RQ1.] What kind of translation quality can we obtain in an extremely low-resource scenario?

\item [CT1.] We have made extensive comparisons with multiple architectures and MT paradigms
to show how difficult the problem is. We have also explored the utility of back-translation and show that it is ineffective given the poor performance of base MT systems used to generate pseudo-parallel data. Our systematic exploration shows that multilingualism is extremely useful for in-domain translation with very limited corpora (see \Sec{baseline}). This type of exhaustive exploration has been missing from most existing works.

\item [RQ2.] What are the effective ways to exploit out-of-domain data for extremely low-resource in-domain translation?

\item [CT2.] Our proposal is to first train a multilingual NMT model on out-of-domain \bidir{Ja}{En} and \bidir{Ru}{En} data, then fine-tune it on in-domain \bidir{Ja}{En} and \bidir{Ru}{En} data, and further fine-tune it on \bidir{Ja}{Ru} data (see \Sec{main}). We show that this stage-wise fine-tuning is crucial for high-quality translation. We then show that the improved NMT models lead to pseudo-parallel data of better quality. This data can then be used to improve the performance even further thereby enabling the generation of better pseudo-parallel data. By iteratively generating pseudo-parallel data and fine-tuning the model on said data, we can achieve the best performance for \bidir{Japanese}{Russian} translation.

\end{description}

To the best of our knowledge, we are the first to perform such an extensive evaluation of extremely low-resource MT problem and propose a novel multilingual multistage fine-tuning approach involving multilingual modeling and domain adaptation to address it.

\section{Our Japanese--Russian Setting}
\label{sec:data}

In this paper, we deal with \bidir{Ja}{Ru} news translation.
This language pair is very challenging because the languages involved have completely different writing system, phonology, morphology, grammar, and syntax.
Among various domains, we experimented with translations in the news domain, considering the importance of sharing news between different language speakers.
Moreover, news domain is one of the most challenging tasks, due to large presence of out-of-vocabulary (OOV) tokens and long sentences.\footnote{News domain translation is also the most competitive tasks in WMT indicating its importance.}
To establish and evaluate existing methods, we also involved English as the third language. As direct parallel corpora are scarce, involving a language such as English for pivoting is quite common \cite{utiyama:07}.

There has been no clean held-out parallel data for \bidir{Ja}{Ru} and \bidir{Ja}{En} news translation.
Therefore, we manually compiled development and test sets using News Commentary data\footnote{\url{http://opus.nlpl.eu/News-Commentary-v11.php}} as a source.
Since the given \bidir{Ja}{Ru} and \bidir{Ja}{En} data share many lines in the Japanese side, we first compiled tri-text data.
Then, from each line, corresponding parts across languages were manually identified, and unaligned parts were split off into a new line.
Note that we have never merged two or more lines.
As a result, we obtained 1,654 lines of data comprising trilingual, bilingual, and monolingual segments (mainly sentences) as summarized in \Tab{nc}.
Finally, for the sake of comparability, we randomly chose 600 trilingual sentences to create a test set, and concatenated the rest of them and bilingual sentences to form development sets.

Our manually aligned development and test sets are publicly available.\footnote{\url{https://github.com/aizhanti/JaRuNC}}

\begin{table}[t]
\centering
\scalebox{.67}{
\begin{tabular}{lll|r|cc}
\hline
\multirow{2}{*}{Ru} &\multirow{2}{*}{Ja} &\multirow{2}{*}{En} & \multirow{2}{*}{\#sent.} &\multicolumn{2}{c}{Usage} \\
& & & &test &development\\
\hline
\checkmark &\checkmark &\checkmark &913 &600 &313 \\
\checkmark &\checkmark &           &173 &-   &173 \\
           &\checkmark &\checkmark &276 &-   &276 \\
\checkmark &           &\checkmark &0   &-   &- \\
\checkmark &           &           &4   &-   &- \\
           &\checkmark &           &287 &-   &- \\
           &           &\checkmark &1   &-   &- \\
\hline
\multicolumn{3}{l|}{Total} &1,654 &- &- \\
\hline
\end{tabular}
}
\caption{Manually aligned News Commentary data.}
\label{tab:nc}
\end{table}

\section{Related Work}
\label{sec:relwork}

\newcite{koehn-knowles:2017:NMT} showed that NMT is unable to handle low-resource language pairs as opposed to PBSMT. Transfer learning approaches \cite{DBLP:journals/corr/FiratCB16:original,DBLP:conf/emnlp/ZophYMK16:original,kocmi-bojar-2018-trivial} work well when a large helping parallel corpus is available. This restricts one of the source or the target languages to be English which, in our case, is not possible. Approaches involving bi-directional NMT modeling is shown to drastically improve low-resource translation \cite{W18-2710}. However, like most other, this work focuses on translation from and into English.

Remaining options include (a) unsupervised MT \cite{D18-1399,D18-1549,marie:usmt-unmt}, (b) parallel sentence mining from non-parallel or comparable corpora \cite{utiyama:03,tillmann:09}, (c) generating pseudo-parallel data \cite{sennrich-haddow-birch:2016:P16-11}, and (d) MT based on pivot languages \cite{utiyama:07}.
The linguistic distance between Japanese and Russian
makes it extremely difficult to learn bilingual knowledge, such as bilingual lexicons and bilingual word embeddings. Unsupervised MT is thus not promising yet, due to its heavy reliance on accurate bilingual word embeddings.  Neither does parallel sentence mining, due to the difficulty of obtaining accurate bilingual lexicons.
Pseudo-parallel data
can be used to augment existing parallel corpora for training, and previous work has reported that such data generated by so-called back-translation can substantially improve the quality of NMT.  However, this approach requires base MT systems that can generate somewhat accurate translations. It is thus infeasible in our scenario, because we can obtain only a weak system which is the consequence of an extremely low-resource situation.
MT based on pivot languages requires large in-domain parallel corpora involving the pivot languages. This technique is thus infeasible, because the in-domain parallel corpora for \bidir{Ja}{En} and \bidir{Ru}{En} pairs are also extremely limited, whereas there are large parallel corpora in other domains.
\Sec{baseline} empirically confirms the limit of these existing approaches.

Fortunately, there are two useful transfer learning solutions using NMT: (e) multilingual modeling to incorporate multiple language pairs into a single model \cite{TACL1081} and (f) domain adaptation to incorporate out-of-domain data \cite{chu:2017:ACL:original}. In this paper, we explore a novel method involving step-wise fine-tuning to combine these two methods. By improving the translation quality in this way, we can also increase the likelihood of pseudo-parallel data being useful to further improve translation quality.

\section{Limit of Using only In-domain Data}
\label{sec:baseline}

This section answers our first research question, [RQ1], about the translation quality that we can achieve using existing methods and in-domain parallel and monolingual data.
We then use the strongest model
to conduct experiments on generating and utilizing back-translated pseudo-parallel data for augmenting NMT. Our intention is to empirically identify the most effective practices as well as recognize the limitations of relying only on in-domain parallel corpora.

\begin{table}[t]
\centering
\scalebox{.67}{
\begin{tabular}{l|c|r|c|c}
\hline 
Lang.pair &Partition &\multicolumn{1}{c|}{\#sent.} &\#tokens &\#types \\
\hline

\multirow{3}{*}{\bidir{Ja}{Ru}}
&train   & 12,356 & 341k / 229k & 22k / 42k \\ 
&development     & 486    & 16k / 11k   & 2.9k / 4.3k \\ 
&test    & 600    & 22k / 15k   & 3.5k / 5.6k \\ 
\hline

\multirow{3}{*}{\bidir{Ja}{En}}
&train   & 47,082 & 1.27M / 1.01M & 48k / 55k \\ 
&development     & 589    & 21k / 16k     & 3.5k / 3.8k \\ 
&test    & 600    & 22k / 17k     & 3.5k / 3.8k \\ 
\hline

\multirow{3}{*}{\bidir{Ru}{En}}
&train   & 82,072 & 1.61M / 1.83M & 144k / 74k \\ 
&development     & 313  & 7.8k / 8.4k     & 3.2k / 2.3k \\ 
&test    & 600  & 15k / 17k     & 5.6k / 3.8k \\ 
\hline
\end{tabular}
}
\caption{Statistics on our in-domain parallel data.}
\label{tab:parallel_data}
\end{table}

\subsection{Data}

To train MT systems among the three languages, i.e., Japanese, Russian, and English, we used all the parallel data provided by Global Voices,\footnote{\url{https://globalvoices.org/}} more specifically those available at OPUS.\footnote{\url{http://opus.nlpl.eu/GlobalVoices-v2015.php}} 
\Tab{parallel_data} summarizes the size of train/development/test splits used in our experiments.
The number of parallel sentences for \bidir{Ja}{Ru} is 12k, for \bidir{Ja}{En} is 47k, and for \bidir{Ru}{En} is 82k.
Note that the three corpora are not mutually exclusive: 9k out of 12k sentences in the \bidir{Ja}{Ru} corpus were also included in the other two parallel corpora, associated with identical English translations. This puts a limit on the positive impact that the helping corpora can have on the translation quality.

Even when one focuses on low-resource language pairs, we often have access to larger quantities of in-domain monolingual data of each language.
Such monolingual data are useful to improve quality of MT, for example, as the source of pseudo-parallel data for augmenting training data for NMT \cite{sennrich-haddow-birch:2016:P16-11} and as the training data for large and smoothed language models for PBSMT \cite{koehn-knowles:2017:NMT}.
\Tab{monolingual_data} summarizes the statistics on our monolingual corpora for several domains including the news domain. Note that we removed from the Global Voices monolingual corpora those sentences that are already present in the parallel corpus.

\begin{table}[t]
\centering
\scalebox{.67}{
\begin{tabular}{l|rrr}
\hline
Corpus        & \multicolumn{1}{c}{Ja} & \multicolumn{1}{c}{Ru} & \multicolumn{1}{c}{En}\\ \hline
Global Voices\footnotemark[5]      & 26k  & 24k  & 842k \\
Wikinews\footnotemark              & 37k  & 243k & -    \\
News Crawl\footnotemark            & -    & 72M  & 194M \\
Yomiuri (2007--2011)\footnotemark  & 19M  & -    & -    \\ \hline
IWSLT\footnotemark                 & 411k & 64k  & 66k  \\
Tatoeba\footnotemark               & 5k   & 58k  & 208k \\ \hline
\end{tabular}
}
\caption{Number of lines in our monolingual data. Whereas the first four are from the news corpora (in-domain), the last two, i.e., ``IWSLT'' and ``Tatoeba,'' are from other domains.}
\label{tab:monolingual_data}
\end{table}
\addtocounter{footnote}{-5}
\stepcounter{footnote}\footnotetext{\url{https://dumps.wikimedia.org/backup-index.html} (20180501)}
\stepcounter{footnote}\footnotetext{\url{http://www.statmt.org/wmt18/translation-task.html}}
\stepcounter{footnote}\footnotetext{\url{https://www.yomiuri.co.jp/database/glossary/}}
\stepcounter{footnote}\footnotetext{\url{http://www.cs.jhu.edu/~kevinduh/a/multitarget-tedtalks/}}
\stepcounter{footnote}\footnotetext{\url{http://opus.nlpl.eu/Tatoeba-v2.php}}

We tokenized English and Russian sentences using \textit{tokenizer.perl} of \texttt{Moses} \cite{koehn-EtAl:2007:PosterDemo}.\footnote{\url{https://github.com/moses-smt/mosesdecoder}}
To tokenize Japanese sentences, we used \texttt{MeCab}\footnote{\url{http://taku910.github.io/mecab}, version 0.996.} with the IPA dictionary.
After tokenization, we eliminated duplicated sentence pairs and sentences with more than 100 tokens for all the languages.

\subsection{MT Methods Examined}

We began with evaluating standard MT paradigms, i.e., PBSMT \cite{koehn-EtAl:2007:PosterDemo} and NMT \cite{DBLP:journals/corr/SutskeverVL14}.  As for PBSMT, we also examined two advanced methods: pivot-based translation relying on a helping language \cite{utiyama:07} and induction of phrase tables from monolingual data \cite{marie:usmt-unmt}.

As for NMT, we compared two types of encoder-decoder architectures: attentional RNN-based model (RNMT) \cite{DBLP:journals/corr/BahdanauCB14:original} and the Transformer model \cite{NIPS2017_7181}.
In addition to standard uni-directional modeling, to cope with the low-resource problem, we examined two multi-directional models: bi-directional model \cite{W18-2710} and multi-to-multi (M2M) model \cite{TACL1081}.

After identifying the best model, we also examined the usefulness of a data augmentation method based on back-translation \cite{sennrich-haddow-birch:2016:P16-11}.

\subsubsection*{PBSMT Systems}

First, we built a PBSMT system for each of the six translation directions.
We obtained phrase tables from parallel corpus using \texttt{SyMGIZA++}\footnote{\url{https://github.com/emjotde/symgiza-pp}} with the \texttt{grow-diag-final} heuristics for word alignment, and \texttt{Moses} for phrase pair extraction.
Then, we trained a bi-directional MSD (monotone, swap, and discontinuous) lexicalized reordering model.
We also trained three 5-gram language models, using \texttt{KenLM}\footnote{\url{https://github.com/kpu/kenlm}} on the following monolingual data: (1) the target side of the parallel data, (2) the concatenation of (1) and the monolingual data from Global Voices, and (3) the concatenation of (1) and all monolingual data in the news domain in \Tab{monolingual_data}.

Subsequently, using English as the pivot language, we examined the following three types of pivot-based PBSMT systems \cite{utiyama:07,cohn:07} for each of \unidir{Ja}{Ru} and \unidir{Ru}{Ja}.
\begin{description}\itemsep=0mm
\item[Cascade:] 2-step decoding using the source-to-English and English-to-target systems.
\item[Synthesize:] Obtain a new phrase table from synthetic parallel data generated by translating English side of the target--English training parallel data to the source language with the English-to-source system.
\item[Triangulate:] Compile a new phrase table combining those for the source-to-English and English-to-target systems.
\end{description}
Among these three, triangulation is the most computationally expensive method. Although we had filtered the component phrase tables using the statistical significance pruning method \cite{D07-1103}, triangulation can generate an enormous number of phrase pairs.
To reduce the computational cost during decoding and the negative effects of potentially noisy phrase pairs, we retained for each source phrase $s$ only the $k$-best translations $t$ according to the forward translation probability $\phi(t|s)$ calculated
from the conditional probabilities in the component models as defined in \newcite{utiyama:07}.
For each of the retained phrase pairs, we also calculated the backward translation probability, $\phi(s|t)$, and lexical translation probabilities, $\phi_{\mathit{lex}}(t|s)$ and $\phi_{\mathit{lex}}(s|t)$, in the same manner as $\phi(t|s)$.

\begin{table*}[t]
\centering
\scalebox{.67}{
\begin{tabular}{c|l|c|c|c|c|c}
\hline
\multirow{2}{*}{ID} & \multirow{2}{*}{System} & \multicolumn{3}{c|}{Parallel data}
  & Total size of & Vocabulary \\
& & \bidir{Ja}{Ru} & \bidir{Ja}{En} & \bidir{Ru}{En} &training data &size \\
\hline
\multirow{3}{*}{(a1), (b1)}
& \unidir{Ja}{Ru} or \unidir{Ru}{Ja} & 12k & -   & -   & 12k & 16k  \\
& \unidir{Ja}{En} or \unidir{En}{Ja} & -   & 47k & -   & 47k & 16k  \\
& \unidir{Ru}{En} or \unidir{En}{Ru} & -   & -   & 82k & 82k & 16k  \\
\hline
\multirow{3}{*}{(a2), (b2)}
& \unidir{Ja}{Ru} and \unidir{Ru}{Ja} & 12k & -   & -   & 24k  & 16k  \\
& \unidir{Ja}{En} and \unidir{En}{Ja} & -   & 47k & -   & 94k  & 16k  \\
& \unidir{Ru}{En} and \unidir{En}{Ru} & -   & -   & 82k & 164k & 16k  \\
\hline
(a3), (b3) &  M2M systems & \unidir{12k}{82k} & \unidir{47k}{82k} & 82k & 492k & 32k \\
\hline
\end{tabular}
}
\caption{Configuration of uni-, bi-directional, and M2M NMT baseline systems. Arrows in ``Parallel data'' columns indicate the over-sampling of the parallel data to match the size of the largest parallel data.}
\label{tab:configuration}
\end{table*}

We also investigated the utility of recent advances in unsupervised MT.
Even though we began with a publicly available implementation of unsupervised PBSMT \cite{D18-1549},\footnote{\url{https://github.com/facebookresearch/UnsupervisedMT}} it crashed due to unknown reasons.
We therefore followed another method described in \newcite{marie:usmt-unmt}.
Instead of short $n$-grams \cite{D18-1399,D18-1549}, we collected a set of phrases in Japanese and Russian from respective monolingual data using the \texttt{word2phrase} algorithm \cite{Mikolov:2013:DRW:2999792.2999959},\footnote{\url{https://code.google.com/archive/p/word2vec/}} as in \newcite{marie:usmt-unmt}.  
To reduce the complexity, we used randomly selected 10M monolingual sentences, and 300k most frequent phrases made of words among the 300k most frequent words.  For each source phrase $s$, we selected 300-best target phrases $t$ according to the translation probability as in \newcite{D18-1549}:
$p(t|s)=\frac{\exp(\beta\cos(\mathit{emb}(t),\mathit{emb}(s))}{\sum_{t'}\exp(\beta\cos(\mathit{emb}(t'),\mathit{emb}(s))},$
where $\mathit{emb}(\cdot)$ stands for a bilingual embedding of a given phrase, obtained through averaging bilingual embeddings of constituent words learned from the two monolingual data using \texttt{fastText}\footnote{\url{https://fasttext.cc/}} and \texttt{vecmap}.\footnote{\url{https://github.com/artetxem/vecmap}}
For each of the retained phrase pair, $p(s|t)$ was computed analogously.  We also computed lexical translation probabilities relying on those learned from the given small parallel corpus.

Up to four phrase tables were jointly exploited by the multiple decoding path ability of \texttt{Moses}.
Weights for the features were tuned using \texttt{KB-MIRA} \cite{cherry:12:a} on the development set; we took the best weights after 15 iterations.
Two hyper-parameters, namely, $k$ for the number of pivot-based phrase pairs per source phrase and $d$ for distortion limit, were determined by a grid search on $k\in\{10,20,40,60,80,100\}$ and $d\in\{8,10,12,14,16,18,20\}$.  In contrast, we used predetermined hyper-parameters for phrase table induction from monolingual data, following the convention: 200 for the dimension of word and phrase embeddings and $\beta=30$.

\begin{table*}[t]
\centering
\scalebox{.67}{
\begin{tabular}{c|l|r|r|r|r|r|r}
\hline
ID & System & \unidir{Ja}{Ru} & \unidir{Ru}{Ja} & \unidir{Ja}{En} & \unidir{En}{Ja} & \unidir{Ru}{En} & \unidir{En}{Ru} \\
\hline
(a1) & Uni-directional RNMT & 0.58 & 1.86 & 2.41 & 7.83 & 18.42 & 13.64 \\
(a2) & Bi-directional RNMT  & 0.65 & 1.61 & 6.18 & 8.81 & 19.60 & 15.11 \\
(a3) & M2M RNMT             & 1.51 & 4.29 & 5.15 & 7.55 & 14.24 & 10.86 \\
\hline
(b1) & Uni-directional Transformer & 0.70     & 1.96     & 4.36      & 7.97      & 20.70 & 16.24 \\
(b2) & Bi-directional Transformer  & 0.19     & 0.87     & 6.48      & 10.63     & 22.25 & 16.03 \\
(b3) & M2M Transformer             & \bf 3.72 & \bf 8.35 & \bf 10.24 & \bf 12.43 & 22.10 & \bf 16.92 \\
\hline
(c1) & Uni-directional supervised PBSMT & 2.02 & 4.45 & 8.19 & 10.27 & \bf 22.37 & 16.52 \\
\hline
\end{tabular}
}
\caption{BLEU scores of baseline systems. \textbf{Bold} indicates the best BLEU score for each translation direction.}
\label{tab:baselines}
\end{table*}

\subsubsection*{NMT Systems}

We used the open-source implementation of the RNMT and the Transformer models in \texttt{tensor2tensor}.\footnote{\url{https://github.com/tensorflow/tensor2tensor}, version 1.6.6.}
A uni-directional model for each of the six translation directions was trained on the corresponding parallel corpus. Bi-directional and M2M models were realized by adding an artificial token that specifies the target language to the beginning of each source sentence and shuffling the entire training data \cite{TACL1081}.

\Tab{configuration} contains some specific hyper-parameters\footnote{We compared two mini-batch sizes, 1024 and 6144 tokens, and found that 6144 and 1024 worked better for RNMT and Transformer, respectively.} for our baseline NMT models.
The hyper-parameters not mentioned in this table used the default values in \texttt{tensor2tensor}.
For M2M systems, we over-sampled \unidir{Ja}{Ru} and \unidir{Ja}{En} training data so that their sizes match the largest \unidir{Ru}{En} data.
To reduce the number of unknown words, we used \texttt{tensor2tensor}'s internal sub-word segmentation mechanism. Since we work in a low-resource setting, we used shared sub-word vocabularies of size 16k for the uni- and bi-directional models and 32k for the M2M models.
The number of training iterations was determined by early-stopping: we evaluated our models on the development set every 1,000 updates, and stopped training if BLEU score for the development set was not improved for 10,000 updates (10 check-points). Note that the development set was created by concatenating those for the individual translation directions without any over-sampling.

Having trained the models, we averaged the last 10 check-points and decoded the test sets with a beam size of 4 and a length penalty which was tuned by a linear search on the BLEU score for the development set.

Similarly to PBSMT, we also evaluated ``Cascade'' and ``Synthesize'' methods with uni-directional NMT models.

\subsection{Results}

We evaluated MT models using case-sensitive and tokenized BLEU \cite{Papineni:2002:BMA:1073083.1073135} on test sets, using \texttt{Moses}'s \textit{multi-bleu.perl}.
Statistical significance ($p<0.05$) on the difference of BLEU scores was tested by \texttt{Moses}'s \textit{bootstrap-hypothesis-difference-significance.pl}.

Tables \ref{tab:baselines} and \ref{tab:pivot} show BLEU scores of all the models, except the NMT systems augmented with back-translations.  Whereas some models achieved reasonable BLEU scores for \bidir{Ja}{En} and \bidir{Ru}{En} translation, all the results for \bidir{Ja}{Ru}, which is our main concern, were abysmal.

Among the NMT models, Transformer models (b$\ast$) were proven to be better than RNMT models (a$\ast$).  RNMT models could not even outperform the uni-directional PBSMT models (c1).  M2M models (a3) and (b3) outperformed their corresponding uni- and bi-directional models in most cases.
It is worth noting that in this extremely low-resource scenario, BLEU scores of the M2M RNMT model for the largest language pair, i.e., \bidir{Ru}{En}, were lower than those of the uni- and bi-directional RNMT models as in \newcite{TACL1081}.
In contrast, with the M2M Transformer model, \bidir{Ru}{En} also benefited from multilingualism. 

Standard PBSMT models (c1) achieved higher BLEU scores than uni-directional NMT models (a1) and (b1), as reported by \newcite{koehn-knowles:2017:NMT}, whereas they underperform the M2M Transformer NMT model (b3).
As shown in \Tab{pivot}, pivot-based PBSMT systems always achieved higher BLEU scores than (c1).  The best model with three phrase tables, labeled ``Synthesize / Triangulate / Gold,'' brought visible BLEU gains with substantial reduction of OOV tokens (\unidir{3047}{1180} for \unidir{Ja}{Ru}, \unidir{4463}{1812} for \unidir{Ru}{Ja}).  However, further extension with phrase tables induced from monolingual data did not push the limit, despite their high coverage; only 336 and 677 OOV tokens were left for the two translation directions, respectively.
This is due to the poor quality of the bilingual word embeddings used to extract the phrase table, as envisaged in \Sec{relwork}.

None of pivot-based approaches with uni-directional NMT models could even remotely rival the M2M Transformer NMT model (b3). 

\begin{table}[t]
\centering
\scalebox{.67}{
\begin{tabular}{l|r|r}\hline
System &\unidir{Ja}{Ru} &\unidir{Ru}{Ja}\\
\hline
PBSMT: Cascade &3.65 &7.62\\
PBSMT: Synthesize &3.37 &6.72\\
PBSMT: Synthesize / Gold &2.94 &6.95\\
PBSMT: Synthesize + Gold &3.07 &6.62\\
PBSMT: Triangulate &\bf 3.75 &7.02\\
PBSMT: Triangulate / Gold &\bf 3.93 &7.02\\
PBSMT: Synthesize / Triangulate / Gold &\bf 4.02 &7.07\\
\hline
PBSMT: Induced &0.37 &0.65\\
PBSMT: Induced / Synthesize / Triangulate / Gold &2.85 &6.86\\
\hline
RNMT: Cascade &1.19 &6.73\\
RNMT: Synthesize &1.82 &3.02\\
RNMT: Synthesize + Gold & 1.62 &3.24 \\
Transformer NMT: Cascade &2.41 &6.84\\
Transformer NMT: Synthesize &1.78 &5.43\\
Transformer NMT: Synthesize + Gold &2.13 &5.06\\
\hline
\end{tabular}
}
\caption{BLEU scores of pivot-based systems. ``Gold'' refers to the phrase table trained on the parallel data. \textbf{Bold} indicates the BLEU score higher than the best one in \Tab{baselines}. ``/'' indicates the use of separately trained multiple phrase tables, whereas so does ``+'' training on the mixture of parallel data.}
\label{tab:pivot}
\end{table}

\begin{table*}[t]
\centering
\scalebox{.67}{
\begin{tabular}{c|l|c|l|l|l|c}
\hline
\multirow{2}{*}{ID} & \multirow{2}{*}{System} & \multicolumn{4}{c|}{Parallel data} & Total size of \\
& &Pseudo & \multicolumn{1}{c|}{\bidir{Ja}{Ru}} & \multicolumn{1}{c|}{\bidir{Ja}{En}} & \multicolumn{1}{c|}{\bidir{Ru}{En}} &training data \\
\hline
\multirow{4}{*}{\#1--\#10}
& \unidir{Ja$\ast$}{Ru} and/or \unidir{Ru$\ast$}{Ja} & \unidir{12k}{82k} & \unidir{12k}{82k}          & \unidir{47k}{82k$\times$2} & 82k$\times$2 & 984k \\
& \unidir{Ja$\ast$}{En} and/or \unidir{En$\ast$}{Ja} & \unidir{47k}{82k} & \unidir{12k}{82k$\times$2} & \unidir{47k}{82k}          & 82k$\times$2 & 984k \\
& \unidir{Ru$\ast$}{En} and/or \unidir{En$\ast$}{Ru} & 82k               & \unidir{12k}{82k$\times$2} & \unidir{47k}{82k$\times$2} & 82k          & 984k \\
& All                                    	     & All of the above  & \unidir{12k}{82k}          & \unidir{47k}{82k}          & 82k          & 984k \\
\hline
\end{tabular}
}
\caption{Over-sampling criteria for pseudo-parallel data generated by back-translation.}
\label{tab:oversampling}
\end{table*}

\begin{table*}[t]
\centering
\scalebox{.67}{
\begin{tabular}{c|c|c|c|c|c|c|r|r|r|r|r|r}
\hline
\multirow{2}{*}{ID} &\multicolumn{6}{c|}{Pseudo-parallel data involved} &\multicolumn{6}{c}{BLEU score} \\
\cline{2-13}
& \unidir{Ja$\ast$}{Ru} & \unidir{Ru$\ast$}{Ja} & \unidir{Ja$\ast$}{En} & \unidir{En$\ast$}{Ja} & \unidir{Ru$\ast$}{En} & \unidir{En$\ast$}{Ru} & \unidir{Ja}{Ru} & \unidir{Ru}{Ja} & \unidir{Ja}{En} & \unidir{En}{Ja} & \unidir{Ru}{En} & \unidir{En}{Ru} \\
\hline
(b3) &- &- &- &- &- &- & 3.72 & 8.35 & 10.24 & 12.43 & 22.10 & 16.92 \\
\hline
\#1 &\checkmark &- &- &- &- &-          & {\winM}\bf 4.59 & \bf 8.63        & \bf 10.64 & \bf 12.94        & \bf 22.21 & \bf 17.30 \\
\#2 &- &\checkmark &- &- &- &-          & \bf 3.74        & {\winM}\bf 8.85 & 10.13     & \bf 13.05        & \bf 22.48 & \bf 17.20 \\
\#3 &\checkmark &\checkmark &- &- &- &- & {\winM}\bf 4.56 & {\winM}\bf 9.09 & \bf 10.57 & {\winM}\bf 13.23 & \bf 22.48 & {\winM}\bf 17.89 \\
\cdashline{2-13}
\#4 &- &- &\checkmark &- &- &-          & 3.71 & 8.05 & {\winM}\bf 11.00 & \bf 12.66        & \bf 22.17 & 16.76 \\
\#5 &- &- &- &\checkmark &- &-          & 3.62 & 8.10 & 9.92             & {\winM}\bf 14.06 & 21.66     & 16.68 \\
\#6 &- &- &\checkmark &\checkmark &- &- & 3.61 & 7.94 & {\winM}\bf 11.51 & {\winM}\bf 14.38 & \bf 22.22	& 16.80 \\
\cdashline{2-13}
\#7 &- &- &- &- &\checkmark &-          & \bf 3.80 & \bf 8.37 & \bf 10.67 & \bf 13.00 & \bf 22.51        & {\winM}\bf 17.73 \\
\#8 &- &- &- &- &- &\checkmark          & \bf 3.77 & 8.04     & \bf 10.52 & 12.43     & {\winM}\bf 22.85 & \bf 17.13 \\
\#9 &- &- &- &- &\checkmark &\checkmark & 3.37     & 8.03     & 10.19     & \bf 12.79 & \bf 22.77        & \bf 17.26 \\
\cdashline{2-13}
\#10 &\checkmark &\checkmark &\checkmark &\checkmark &\checkmark &\checkmark & {\winM}\bf 4.43 & {\winM}\bf 9.38 & {\winM}\bf 12.06 & {\winM}\bf 14.43 & {\winM}\bf 23.09 & \bf 17.30 \\
\hline
\end{tabular}
}
\caption{BLEU scores of M2M Transformer NMT systems trained on the mixture of given parallel corpus and pseudo-parallel data generated by back-translation using (b3). Six ``\unidir{X$\ast$}{Y}'' columns show whether the pseudo-parallel data for each translation direction is involved.
{\bf Bold} indicates the scores higher than (b3) and ``{\winM}'' indicates statistical significance of the improvement.}
\label{tab:ablation}
\end{table*}

\subsection{Augmentation with Back-translation}
\label{sec:backtrans1}

Given that the M2M Transformer NMT model (b3) achieved best results for most of the translation directions and competitive results for the rest, we further explored it through back-translation.

We examined the utility of pseudo-parallel data for all the six translation directions, unlike the work of \newcite{lakew2017improving} and \newcite{lakew2018comparison}, which concentrate only on the zero-shot language pair, and the work of \newcite{W18-2710}, which compares only uni- or bi-directional models.  We investigated whether each translation direction in M2M models will benefit from pseudo-parallel data and if so, what kind of improvement takes place.

First, we selected sentences to be back-translated from in-domain monolingual data (\Tab{monolingual_data}), relying on the score proposed by \newcite{moore:intelligent} via the following procedure.
\begin{enumerate}\itemsep=0mm
\item For each language, train two 4-gram language models, using KenLM: an in-domain one on all the Global Voices data, i.e., both parallel and monolingual data, and a general-domain one on the concatenation of Global Voices, IWSLT, and Tatoeba data.
\item For each language, discard sentences containing OOVs according to the in-domain language model.
\item For each translation direction, select the $T$-best monolingual sentences in the news domain, according to the difference between cross-entropy scores given by the in-domain and general-domain language models. 
\end{enumerate}

Whereas \newcite{W18-2710} exploited monolingual data much larger than parallel data, we maintained a 1:1 ratio between them \cite{TACL1081}, setting $T$ to the number of lines of parallel data of given language pair.

Selected monolingual sentences were then translated using the M2M Transformer NMT model (b3) to compose pseudo-parallel data.  Then, the pseudo-parallel data were enlarged by over-sampling as summarized in \Tab{oversampling}.  Finally, new NMT models were trained on the concatenation of the original parallel and pseudo-parallel data from scratch in the same manner as the previous NMT models with the same hyper-parameters.

\Tab{ablation} shows the BLEU scores achieved by several reasonable combinations of six-way pseudo-parallel data.
We observed that the use of all six-way pseudo-parallel data (\#10) significantly improved the base model for all the translation directions, except \unidir{En}{Ru}.  A translation direction often benefited when the pseudo-parallel data for that specific direction was used.

\subsection{Summary}

We have evaluated an extensive variation of MT models\footnote{Other conceivable options include transfer learning using parallel data between English and one of Japanese and Russian as either source or target language, such as pre-training an \unidir{En}{Ru} model and fine-tuning it for \unidir{Ja}{Ru}.  Our M2M models conceptually subsume them, even though they do not explicitly divide the two steps during training. On the other hand, our method proposed in \Sec{main} explicitly conducts transfer learning for domain adaptation followed by additional transfer learning across different languages.} that rely only on in-domain parallel and monolingual data.  However, the resulting BLEU scores for \unidir{Ja}{Ru} and \unidir{Ru}{Ja} tasks do not exceed 10 BLEU points, implying the inherent limitation of the in-domain data as well as the difficulty of these translation directions.

\section{Exploiting Large Out-of-Domain Data Involving a Helping Language}
\label{sec:main}

The limitation of relying only on in-domain data demonstrated in \Sec{baseline} motivates us to explore other types of parallel data.  As raised in our second research question, [RQ2], we considered the effective ways to exploit out-of-domain data.

According to language pair and domain, parallel data can be classified into four categories in \Tab{variation}.
Among all the categories, out-of-domain data for the language pair of interest have been exploited in the domain adaptation scenarios (\unidir{C}{A}) \cite{chu:2017:ACL:original}. However, for \bidir{Ja}{Ru}, no out-of-domain data is available.
To exploit out-of-domain parallel data for \bidir{Ja}{En} and \bidir{Ru}{En} pairs instead, we propose a multistage fine-tuning method, which combines two types of transfer learning, i.e., domain adaptation for \bidir{Ja}{En} and \bidir{Ru}{En} (\unidir{D}{B}) and multilingual transfer (\unidir{B}{A}), relying on the M2M model examined in \Sec{baseline}.
We also examined the utility of fine-tuning for iteratively generating and using pseudo-parallel data.

\begin{table}[t]
\centering
\scalebox{.67}{
\begin{tabular}{c|cc}
\hline
    Domain $\backslash$ language pair & Direct & One-side shared \\
\hline
    in-domain & A, $\checkmark$ & B, $\checkmark$ \\
    out-of-domain & C, $\times$ & D, $\checkmark$ \\
\hline
\end{tabular}
}
\caption{Classification of parallel data.}
\label{tab:variation}
\end{table}

\subsection{Multistage Fine-tuning}
\label{sec:mft}

Simply using NMT systems trained on out-of-domain data for in-domain translation is known to perform badly.
In order to effectively use large-scale out-of-domain data for our extremely low-resource task, we propose to perform domain adaptation through either (a) conventional fine-tuning, where an NMT system trained on out-of-domain data is fine-tuned only on in-domain data, or (b) mixed fine-tuning \cite{chu:2017:ACL:original}, where pre-trained out-of-domain NMT system is fine-tuned using a mixture of in-domain and out-of-domain data.
The same options are available for transferring from \bidir{Ja}{En} and \bidir{Ru}{En} to \bidir{Ja}{Ru}.

We inevitably involve two types of transfer learning, i.e., domain adaptation for \bidir{Ja}{En} and \bidir{Ru}{En} and multilingual transfer for \bidir{Ja}{Ru} pair.  Among several conceivable options for managing these two problems, we examined the following multistage fine-tuning.
\begin{description}\itemsep=0mm
    \item[Stage 0. Out-of-domain pre-training:] Pre-train a multilingual model only on the \bidir{Ja}{En} and \bidir{Ru}{En} out-of-domain parallel data (\RNum{1}), where the vocabulary of the model is determined on the basis of the in-domain parallel data in the same manner as the M2M NMT models examined in \Sec{baseline}.
    \item[Stage 1. Fine-tuning for domain adaptation:] Fine-tune the pre-trained model (\RNum{1}) on the in-domain \bidir{Ja}{En} and \bidir{Ru}{En} parallel data (fine-tuning, \RNum{2}) or on the mixture of in-domain and out-of-domain \bidir{Ja}{En} and \bidir{Ru}{En} parallel data (mixed fine-tuning, \RNum{3}). 
    \item[Stage 2. Fine-tuning for \bidir{Ja}{Ru} pair:] Further fine-tune the models (each of \RNum{2} and \RNum{3}) for \bidir{Ja}{Ru} on in-domain parallel data for this language pair only (fine-tuning, \RNum{4} and \RNum{6}) or on all the in-domain parallel data (mixed fine-tuning, \RNum{5} and \RNum{7}).
\end{description}

We chose this way due to the following two reasons.  First, we need to take a balance between several different parallel corpora sizes.
The other reason is division of labor; we assume that solving each sub-problem one by one should enable gradual shift of parameters.

\begin{table}[t]
\centering
\scalebox{.67}{
\begin{tabular}{l|c|r|c|c}
\hline 
Lang.pair & Corpus &\multicolumn{1}{c|}{\#sent.} &\#tokens &\#types\\
\hline
\multirow{1}{*}{\bidir{Ja}{En}}
& ASPEC     & 1,500,000 & 42.3M / 34.6M & 234k / 1.02M \\
\hline
\multirow{2}{*}{\bidir{Ru}{En}}
& UN        & 2,647,243 & 90.5M / 92.8M & 757k / 593k \\
& Yandex    & 320,325   & 8.51M / 9.26M & 617k / 407k \\
\hline
\end{tabular}
}
\caption{Statistics on our out-of-domain parallel data.}
\label{tab:external}
\end{table}

\begin{table*}[t]
\centering
\scalebox{.67}{
\begin{tabular}{c|c|c|c|c|c|c|r|r|r|r|r|r}
\hline
\multirow{2}{*}{ID} & \multirow{2}{*}{Initialized} & \multicolumn{2}{c|}{Out-of-domain data} &\multicolumn{3}{c|}{In-domain data} &\multicolumn{6}{c}{BLEU score} \\ \cline{3-13}
& & \bidir{Ja}{En} & \bidir{Ru}{En} & \bidir{Ja}{Ru} & \bidir{Ja}{En} & \bidir{Ru}{En} & \unidir{Ja}{Ru} & \unidir{Ru}{Ja} & \unidir{Ja}{En} & \unidir{En}{Ja} & \unidir{Ru}{En} & \unidir{En}{Ru} \\
\hline
(b3) &- &- &- &\checkmark &\checkmark &\checkmark     & 3.72  & 8.35 & 10.24 & 12.43 & 22.10 & 16.92 \\
\hline
\RNum{1} &-         &\checkmark &\checkmark &-&-&-                    & 0.00        & 0.15         & 4.59             & 4.15             & {\winM}25.22     & {\winM}20.37 \\
\hline
\RNum{2} & \RNum{1} &-&-&-&\checkmark &\checkmark                     & 0.20        & 0.70         & {\winM}14.10     & {\winM}\bf 17.80 & {\winM}28.23     & {\winM}24.35 \\
\RNum{3} & \RNum{1} &\checkmark &\checkmark &-&\checkmark &\checkmark & 0.23        & 1.07         & {\winM}13.31     & {\winM}17.74     & {\winM}\bf 28.73 & {\winM}\bf 25.22 \\
\hline
\RNum{4} & \RNum{2} &-&-&\checkmark &-&-                              & {\winM}5.44 & {\winM}10.67 & 0.12             & 3.97             & 0.11             & 3.66 \\
\RNum{5} & \RNum{2} &-&-&\checkmark &\checkmark &\checkmark           & {\winM}6.90 & {\winM}11.99 & {\winM}14.34     & {\winM}16.93     & {\winM}27.50     & {\winM}23.17 \\
\hline
\RNum{6} & \RNum{3} &-&-&\checkmark &-&-                              & {\winM}5.91 & {\winM}10.83 & 0.26             & 2.18             & 0.18             & 1.10 \\
\RNum{7} & \RNum{3} &-&-&\checkmark &\checkmark &\checkmark           & {\winM}7.49 & {\winM}12.10 & {\winM}\bf 14.63 & {\winM}17.51     & {\winM}28.51     & {\winM}24.60 \\
\hline\hline
\RNum{1}' &- &\checkmark &\checkmark &\checkmark &\checkmark &\checkmark        & {\winM}5.31     & {\winM}10.73   & {\winM}14.41   & {\winM}16.34 & {\winM}27.46 & {\winM}23.21\\
\RNum{2}' &\RNum{1} &- &- &\checkmark &\checkmark &\checkmark                   & {\winM}6.30     & {\winM}11.64   & {\winM}14.29   & {\winM}16.83 & {\winM}27.53 & {\winM}23.00\\
\RNum{3}' &\RNum{1} &\checkmark &\checkmark &\checkmark &\checkmark &\checkmark & {\winM}\bf 7.53 & {\winM}\bf 12.33 & {\winM}14.19 & {\winM}16.77 & {\winM}27.94 & {\winM}23.97\\
\hline
\end{tabular}
}
\caption{BLEU scores obtained through multistage fine-tuning. ``Initialized'' column indicates the model used for initializing parameters that are fine-tuned on the data indicated by $\checkmark$. \textbf{Bold} indicates the best BLEU score for each translation direction. ``{\winM}'' indicates statistical significance of the improvement over (b3).}
\label{tab:damlftorigindvocab}
\end{table*}

\subsection{Data Selection}

As an additional large-scale out-of-domain parallel data for \bidir{Ja}{En}, we used the cleanest 1.5M sentences from the Asian Scientific Paper Excerpt Corpus (ASPEC) \cite{NAKAZAWA16.621}.\footnote{\url{http://lotus.kuee.kyoto-u.ac.jp/ASPEC/}}
As for \bidir{Ru}{En}, we used the UN and Yandex corpora released for the WMT 2018 News Translation Task.\footnote{\url{http://www.statmt.org/wmt18/translation-task.html}}
We retained \bidir{Ru}{En} sentence pairs that contain at least one OOV token in both sides, according to the in-domain language model trained in \Sec{backtrans1}.
\Tab{external} summarizes the statistics on the remaining out-of-domain parallel data.

\subsection{Results}

\Tab{damlftorigindvocab} shows the results of our multistage fine-tuning, where the IDs of each row refer to those described in \Sec{mft}.
First of all, the final models of our multistage fine-tuning, i.e., \RNum{5} and \RNum{7}, achieved significantly higher BLEU scores than (b3) in \Tab{baselines}, a weak baseline without using any monolingual data, and \#10 in \Tab{ablation}, a strong baseline established with monolingual data.

The performance of the initial model (\RNum{1}) depends on the language pair.  For \bidir{Ja}{Ru} pair, it cannot achieve minimum level of quality since the model has never seen parallel data for this pair.  The performance on \bidir{Ja}{En} pair was much lower than the two baseline models, reflecting the crucial mismatch between training and testing domains.  In contrast, \bidir{Ru}{En} pair benefited the most and achieved surprisingly high BLEU scores.  The reason might be due to the proximity of out-of-domain training data and in-domain test data.

The first fine-tuning stage significantly pushed up the translation quality for \bidir{Ja}{En} and \bidir{Ru}{En} pairs, in both cases with fine-tuning (\RNum{2}) and mixed fine-tuning (\RNum{3}).  At this stage, both models performed only poorly for \bidir{Ja}{Ru} pair as they have not yet seen \bidir{Ja}{Ru} parallel data.  Either model had a consistent advantage to the other.

When these models were further fine-tuned only on the in-domain \bidir{Ja}{Ru} parallel data (\RNum{4} and \RNum{6}), we obtained translations of better quality than the two baselines for \bidir{Ja}{Ru} pair.  However, as a result of complete ignorance of \bidir{Ja}{En} and \bidir{Ru}{En} pairs, the models only produced translations of poor quality for these language pairs.  In contrast, mixed fine-tuning for the second fine-tuning stage (\RNum{5} and \RNum{7}) resulted in consistently better models than conventional fine-tuning (\RNum{4} and \RNum{6}), irrespective of the choice at the first stage, thanks to the gradual shift of parameters realized by in-domain \bidir{Ja}{En} and \bidir{Ru}{En} parallel data.
Unfortunately, the translation quality for \bidir{Ja}{En} and \bidir{Ru}{En} pairs sometimes degraded from \RNum{2} and \RNum{3}.  Nevertheless,  the BLEU scores still retain the large margin against two baselines.

The last three rows in \Tab{damlftorigindvocab} present BLEU scores obtained by the methods with fewer fine-tuning steps.  The most naive model \RNum{1}', trained on the balanced mixture of whole five types of corpora from scratch, and the model \RNum{2}', obtained through a single-step conventional fine-tuning of \RNum{1} on all the in-domain data, achieved only BLEU scores consistently worse than \RNum{7}.
In contrast, when we merged our two fine-tuning steps into a single mixed fine-tuning on \RNum{1}, we obtained a model \RNum{3}' which is better for the \bidir{Ja}{Ru} pair than \RNum{7}.  Nevertheless, they are still comparable to those of \RNum{7} and the BLEU scores for the other two language pairs are much lower than \RNum{7}.  As such, we conclude that our multistage fine-tuning leads to a more robust in-domain multilingual model.

\begin{table*}[t]
\centering
\scalebox{.68}{
\begin{tabular}{c|c|c|r|r|r|r|r|r}
\hline
\multirow{2}{*}{No} & \multirow{2}{*}{Initialized} & \multirow{2}{*}{BT} &\multicolumn{6}{c}{BLEU score} \\ \cline{4-9}
&&& \unidir{Ja}{Ru} & \unidir{Ru}{Ja} & \unidir{Ja}{En} & \unidir{En}{Ja} & \unidir{Ru}{En} & \unidir{En}{Ru} \\
\hline
\#10 &- & (b3) & 4.43 & 9.38 & 12.06 & 14.43 & 23.09 & 17.30 \\
\hline
new \#10 &-& \RNum{7}               & {\winM}6.55 & {\winM}11.36 & {\winM}13.77 & {\winM}15.59 & {\winM}24.91 & {\winM}20.55 \\
\hline
\RNum{8} & \RNum{7}  & \RNum{7}   & {\winM}7.83 & {\winM}12.21 & {\winM}15.06 & {\winM}17.19 & {\winM}28.49 & {\winM}23.96 \\
\RNum{9} & \RNum{8} & \RNum{8}    & {\winM}8.03 & {\winM}12.55 & {\winM}15.07 & {\winM}17.80 & {\winM}28.16 & {\winM}24.27 \\
\RNum{10} & \RNum{9} & \RNum{9}   & {\winM}7.76 & {\winM}12.59 & {\winM}15.08 & {\winM}18.12 & {\winM}28.18 & {\winM}24.67 \\
\RNum{11} & \RNum{10} & \RNum{10} & {\winM}7.85 & {\winM}12.97 & {\winM}15.26 & {\winM}17.83 & {\winM}28.49 & {\winM}24.36 \\
\RNum{12} & \RNum{11} & \RNum{11} & {\winM}8.16 & {\winM}13.09 & {\winM}14.96 & {\winM}17.74 & {\winM}28.45 & {\winM}24.35 \\
\hline
\end{tabular}
}
\caption{BLEU scores achieved through fine-tuning on the mixture of the original parallel data and six-way pseudo-parallel data. ``Initialized'' column indicates the model used for initializing parameters and so does ``BT'' column the model used to generate pseudo-parallel data. ``{\winM}'' indicates statistical significance of the improvement over \#10.}
\label{tab:iterativeBT}
\end{table*}

\subsection{Further Augmentation with Back-translation}
\label{sec:backtrans2}

Having obtained a better model, we examined again the utility of back-translation.
More precisely, we investigated (a) whether the pseudo-parallel data generated by an improved NMT model leads to a further improvement, and (b) whether one more stage of fine-tuning on the mixture of original parallel and pseudo-parallel data will result in a model better than training a new model from scratch as examined in \Sec{backtrans1}.

Given an NMT model, we first generated six-way pseudo-parallel data by translating monolingual data.  For the sake of comparability, we used the identical monolingual sentences sampled in \Sec{backtrans1}.  Then, we further fine-tuned the given model on the mixture of the generated pseudo-parallel data and the original parallel data, following the same over-sampling procedure in \Sec{backtrans1}.  We repeated these steps five times.

\Tab{iterativeBT} shows the results.
``new \#10'' in the second row indicates an M2M Transformer model trained from scratch on the mixture of six-way pseudo-parallel data generated by \RNum{7} and the original parallel data.  It achieved higher BLEU scores than \#10 in \Tab{ablation} thanks to the pseudo-parallel data of better quality, but underperformed the base NMT model \RNum{7}.  In contrast, our fine-tuned model \RNum{8} successfully surpassed \RNum{7}, and one more iteration (\RNum{9}) further improved BLEU scores for all translation directions, except \unidir{Ru}{En}.  
Although further iterations did not necessarily gain BLEU scores, we came to a much higher plateau compared to the results in \Sec{baseline}.

\begin{table}[t]
\centering
\scalebox{.67}{
\begin{tabular}{l|r|r}
\hline
Investigation step & \unidir{Ja}{Ru} & \unidir{Ru}{Ja} \\
\hline
Uni-directional Transformer: (b1) in \Tab{baselines}             & 0.70 & 1.96 \\
M2M Transformer: (b3) in \Tab{baselines}                         & 3.72 & 8.35 \\
+ six-way pseudo-parallel data: \#10 in \Tab{ablation}           & 4.43	& 9.38 \\
\hline
M2M multistage fine-tuning: \RNum{7} in \Tab{damlftorigindvocab} & 7.49	& 12.10 \\
+ six-way pseudo-parallel data: \RNum{12} in \Tab{iterativeBT}   & 8.16 & 13.09 \\
\hline
\end{tabular}
}
\caption{Summary of our investigation: BLEU scores of the best NMT systems at each step.}
\label{tab:summary}
\end{table}

\section{Conclusion}
In this paper, we challenged the difficult task of \bidir{Ja}{Ru} news domain translation in an extremely low-resource setting. We empirically confirmed the limited success of well-established solutions when restricted to in-domain data. Then, to incorporate out-of-domain data, we proposed a multilingual multistage fine-tuning approach and observed that it substantially improves \bidir{Ja}{Ru} translation by over 3.7 BLEU points compared to a strong baseline, as summarized in \Tab{summary}. This paper contains an empirical comparison of several existing approaches and hence we hope that our paper can act as a guideline to researchers attempting to tackle extremely low-resource translation.

In the future, we plan to confirm further fine-tuning for each of specific translation directions.
We will also explore the way to exploit out-of-domain pseudo-parallel data, better domain-adaptation approaches, and additional challenging language pairs.

\section*{Acknowledgments}
This work was carried out when Aizhan Imankulova was taking up an internship at NICT, Japan.
We would like to thank the reviewers for their insightful comments.
A part of this work was conducted under the program ``Promotion of Global Communications Plan: Research, Development, and Social Demonstration of Multilingual Speech Translation Technology'' of the Ministry of Internal Affairs and Communications (MIC), Japan.

\bibliographystyle{acl_natbib}
\bibliography{mtsummit2019}

\end{document}